# Exploiting Cross-Document Relations for Multi-document Evolving Summarization

Stergos D. Afantenos[1], Irene Doura[2],
Eleni Kapellou[2], and Vangelis Karkaletsis[1]

[1] Software and Knowledge Engineering Laboratory
Institute of Informatics and Telecommunications,
National Center for Scientific Research (NCSR) "Demokritos"
{stergos,vangelis}@iit.demokritos.gr

[2] Institute of Language and Speech Processing
intoura@isll.uoa.gr, kotkap@teiath.gr

**Abstract.** This paper presents a methodology for summarization from multiple documents which are about a specific topic. It is based on the specification and identification of the cross-document relations that occur among textual elements within those documents. Our methodology involves the specification of the topic-specific entities, the messages conveyed for the specific entities by certain textual elements and the specification of the relations that can hold among these messages. The above resources are necessary for setting up a specific topic for our query-based summarization approach which uses these resources to identify the query-specific messages within the documents and the query-specific relations that connect these messages across documents.

## 1 Introduction

In the process of reading a text, we come to realize that several textual elements have a sort of connection with other textual elements. That is not a coincidence. Mann and Thompson (1988), in fact, have proposed a theory, the *Rhetorical Structure Theory (RST)*, according to which sentences or phrases are connected with some *relations*, from a set of predefined relations. This theory has been exploited, by Marcu (2000) for example, for single-document summarization.

We do believe that something similar happens across documents, at least when they are on the same topic. In other words, several *"elements"* in one document are *"connected"* with several other *"elements"* in another document. The point, of course, is to define those "elements" and "connections".

The aim of this paper is an attempt to remove the quotes from the words "elements" and "connections", *i.e.* try to make a little bit more explicit what such elements and connections can be, as well as suggest possible ways of how they can be used for multi-document summarization.

The motivation behind this work is presented in the following section, in which the related work will be presented as well. The general methodology of our work, as it has been formed until now, is given in section 3. This methodology is made more explicit through a case study in section 4.

## 2   Related Work – Motivation

As mentioned in the introduction, in this paper we consider the question of whether something similar to the RST can hold for more than one documents, and if that is so, how can that theory be exploited for the automatic creation of summaries. Of course, we are not the only ones who have given this matter some consideration. Radev (2000), inspired by Mann and Thompson's (1988) RST, tried to create a similar theory which would connect multiple documents. He called his theory *Cross-document Structure Theory (CST)*.

In his endeavor Radev (2000) proposed a set of relations which bear a certain similarity to the RST relations, such as *Elaboration, Contradiction, Equivalence, Agreement*, etc[1]. These relations are not applied to phrases or sentences of one document anymore, but, depending on the relation, they can be applied across documents to words, phrases, sentences, paragraphs or even entire documents. Radev claims that Cross-document Structure Theory can be the basis for multi-document summarization.

Since this theory was merely a proposition by Radev, Zhang *et al.* (2002) tried to put that theory to test by conducting an experiment, according to which subjects (judges) were asked to read a set of news articles and write down the CST relations they observed. The set of documents contained 11 news articles that were on the same topic but which originated from different sources. Five pairs of documents were given to 9 subjects, along with instructions on how to annotate the documents with the proposed relations. The instructions contained the set of 24 relations, along with examples on their use. It was clearly mentioned that the set of those relations was simply a "proposed" set and that they should feel free to use their own relations, should they wish to.

The results of this experiment can be summarize them as follows[2]:

- The inter-judge agreement was very low.
- Only a small subset of the proposed relations was used by the judges.
- No new relations were proposed.
- Only sentences were connected with each other; relations between words or phrases or paragraphs and documents were ignored.

We believe that the reasons for these results lie not in the fact that certain "elements" of one document are not "connected" with other "elements" in another document, but in the following.

First of all, the relations that Radev (2000) proposes seem to be similar to the ones that Mann and Thompson (1988) have proposed for the *Rhetorical Structure Theory*, only that they are extended to include multiple documents. While this seems reasonable at first sight, if we delve a little bit more in the details we will see that it is somewhat problematic. RST is based on the assumption of a *coherent text*, whose meaning needs further clarification when extended to include multiple documents written by different authors, under different conditions and in a different context.

---

[1] For more information on the RST relations see Mann and Thompson (1988) or http://www.sil.org/~mannb/rst.

[2] For more details, the reader is encouraged to consult Zhang *et al.* (2002).

A second potential cause for the above results we consider to be the fact that CST concentrates on textual spans, and not — instead — on what these textual spans *represent*. In the context of multiple documents, the connection of textual spans seems more logical if what we connect is what is being *represented* by the textual spans and not the textual spans *themselves*.

The problem, of course, is that in order to find what the textual spans represent, one has to focus on a specific topic. And at this point comes our proposition, that to study the cross-document relations one has to begin with a specific topic, before generalizing, if that is possible.

In the next section we propose a general methodology for the manual specification of the cross-document relations. We also present the architecture of a query-based summarization system that exploits such relations. This system is currently under development. In section 4 we will give a particular example of this methodology in the topic of the description of football matches.

## 3   Methodology for Identifying Cross-Document Relations

The conclusion that can be drawn from the previous section is that the study of general cross-document relations, at least in the sense that Radev (2000) proposes it, is still premature. Instead we propose to concentrate on the identification of the *nature of what* can be connected between the documents first, as well as how that can be connected in each particular topic, and then try to generalize. Before continuing with the presentation of our methodology, we would like to put it in the context of our approach to multi-document summarization. Our approach is a query-based summarization one, which employs: a) an Information Extraction (IE) system for extracting the messages that are needed for the summarization task; this system is used off-line for the processing of the documents before the submission of the query. b) a Natural Language Generation (NLG) system for presenting the summary, exploiting those messages that are relevant to the query within a document and the relations that connect these messages across documents; the NLG system is used on-line after the submission of the query.

We have to stress here that despite the fact that this paper concentrates on the presentation of the methodology and not on the query-based summarization system, which is currently under development. The basic stages of our methodology are presented below.

***Collection of corpus.*** The first stage of our methodology involves the collection of the corpus to be summarized. The corpus should be on a certain topic in which several events — that we want to summarize — are evolving and are being described by more than one source. Although this process can be automated using text classification techniques, we currently do not plan to do so and the collection of the corpus is done manually.

***Creation of a topic-specific ontology.*** The next step involves the specification of the types of entities in the corpus that our summaries will concentrate on, as well as the specification of the events, and the entities' attributes or their roles in those events. For example, in the topic of football matches' descriptions (see the following section), important entity types are team, player, etc; important events are foul, penalty, etc; important entities' roles are the winner team, the player that shot the penalty, etc. In other words we have to set up the topic's *ontology*. The specification of the entity types and the ontology structure is done manually. Yet, there are several ontology editors that enable the building of the ontology using a specific knowledge representation format. In our case, we use the *Protégé*-based ontology editor developed in the CROSSMARC project (Pazienza et al. 2003).

***Specification of the topic-specific message types.*** Our summarization system employs an NLG component which generates text from a set of messages that convey the meaning of the text to be generated (Reiter and Dale 2000). Therefore, the next step in our methodology is to specify the message types in the specific topic where our summarization system will be applied. Those message types should contain entity types and event-specific roles from the ontology that was built beforehand. The specification of the message types and their precise definition results through a study of the corpus.

We consider this step of the methodology as an IE task, which can be performed off-line before the query submission. Once the message types have been specified, the IE sub-system will locate the textual elements which instantiate particular messages and fill in the arguments for each message. For this purpose we use the Greek IE system developed in the context of the CROSSMARC project (Karkaletsis and Spyropoulos 2003), which is currently being adapted to the topics.

***Specification of the topic-specific relations.*** Once we have finalized the set of topic-specific message types that occur within the documents of our corpus, we should try to specify what sort of relation types can connect those messages across the documents, again in relation to our summarization task. The set of relations can be a general one, similar to that of Radev's (2000) CST or Mann and Thompson's (1988) RST, or it can be topic-specific.

In order to define the relations we rely on the message types and the values that they can have in their arguments. Thus, we devise a set of rules to identify the relations connecting the messages. Once the instances of the relations have been identified, the relevant to the query messages and relations can be passed to the NLG system.

The above constitute our general methodology for specifying cross-document relations and exploiting them by a query-based summarization system. Those relations are not a rigid set that will be exactly the same, independently of the summarization task. Instead, they are quite flexible and can be customized for whatever application of summarization one has to deal with. In our case we were interested in the creation of *evolving summaries*, *i.e.* summaries of events within a topic which evolve through time, so our relations are customized for this.

In the next section we will make the above more explicit by presenting our initial steps towards the application of the above methodology to a specific topic: that of the description of football matches.

## 4   Case Study: Descriptions of Football Matches

Our choice for the topic of descriptions of football matches was influenced by the fact that we were interested in the study of *evolving summaries*. In that topic the main events that evolve can easily be isolated from the rest of the events in order to be studied separately, which made it suitable for an initial study.

The target language of this topic was Greek. The first step in the methodology is the collection of the corpus. The corpus we collected originated from three different sources: a newspaper, a sports magazine and the official internet site of a team[3], and it contained descriptions of football matches for the Greek football Championship of the first division for the years 2002–2003. The total number of documents that we studied was 90; they contained about 67,500 tokens totally.

For every team, we organized the descriptions of the matches in a way which reflects a *grid* and is depicted in Fig. 1. Note that if in a particular championship $N$ teams compete, then the total number of rounds will be $(N-1) \times 2$. This grid organization reflects the fact that for a certain team we have two axes in which we can view the descriptions of its matches. The first, *horizontal axis*, contains the descriptions of the *same match* but from *different sources*. The second, *vertical axis*, contains the descriptions from the *same source* but for all the *series* of matches during the championship. It should also be noted that if the grids of the teams are interposed on top of each other, the result will be a *cube* organization for the whole championship.

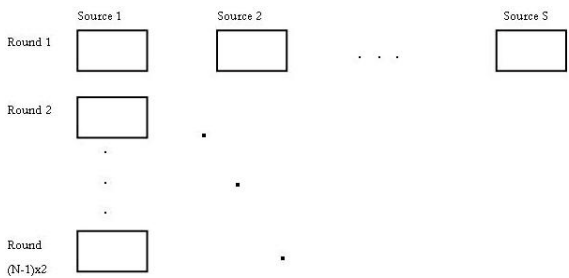

**Fig. 1.** Organization of the corpus in a grid

The next step involves the building of an ontology. Some of the main entities that we have decided to include in the ontology are shown in Fig. 2. Although this ontology is still in its first version and further refinement is still needed, it is in a state that it can be used for our experiments, as we describe below.

---

[3]One could argue that each of our source constitutes a different *genre* of text, since each source has a different target readership and a different purpose.

- TIME
    1. *first half*
    2. *second half*
    3. etc ...
- EVENTS
    1. foul
    2. penalty
    3. corner
    4. etc.
- etc.
- PERSON
    1. *player*
    2. *official*
        (a) *coach*
        (b) *owner*
        (c) etc.
    3. *referees*
    4. *public*
    5. etc.
- TEAM

**Fig. 2.** A high level excerpt of the ontology

This ontology is related to the specification of the message types, which constitutes the next step in our methodology. In Fig. 3 several message types are presented in detail. The complete list of the message types is the following:

**performance, satisfaction, blocks, superior, belongs, final_score, opportunity_lost, change, cancelation_of_goal, surprise, injured, alias, penalty, card, behavior, foul, selection_of_scheme, win, comeback, absent, successive_victories, refereeship, hope_for, scorer, expectations, conditions**

- **performance (entity, in_what, time_span, value)**
  entity    : *TEAM, PERSON*
  in_what   : offense, defense, general etc.
  value     : bad, good, moderate, excellent
  time_span : *TIME*
  Comment   : entity had value performance in_what during time_span
- **satisfaction (entity$_1$, entity$_2$, value)**
  entity$_1$, entity$_2$ : *TEAM, PERSON*
  value              : low ... high
  Comment            : entity$_1$ had value satisfaction from entity$_2$
- **superior (entity$_1$, entity$_2$, in_what, time_span)**
  entity$_1$, entity$_2$ : *TEAM, player*
  in_what            : defense, offense (this argument is optional)
  time_span          : *TIME*
  Comment            : entity$_1$ was superior to entity$_2$ in_what during time_span

**Fig. 3.** Some messages from the football topic

It cannot be claimed that this set of messages is final, since changes in the ontology might result in further refinement of the messages.

The final step of our methodology involves the identification of the relation types that exist between the messages. As it has been noted before, the relations can be similar to Radev's (2000) CST or to Mann and Thompson's (1988) RST, but they can also be different, depicting the needs for summarization that one has. In our case, the *grid* organization of our corpus, along with the fact that we were interested in the study of *evolving summarization*, has led us to create relations across two axes, the *horizontal* and the *vertical* (See Fig. 4).

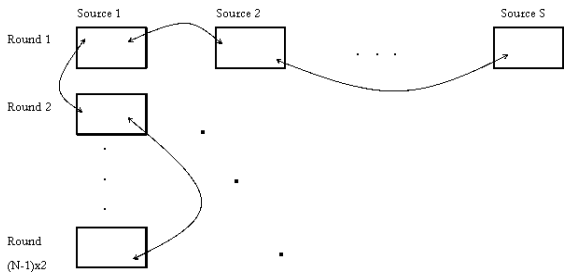

**Fig. 4.** Relations in two axes

Relations on the horizontal axis are concerned with the *same* match as it was described by *different sources*, while relations on the vertical axis are concerned with the *different* matches of a certain team as described by the *same source*. We call the relations on the horizontal axis *synchronic* and the relations on the vertical axis *diachronic*. Those last relations, one could argue, concern the *progress* of a certain team, or of a certain player, which reflects our endeavor for *evolving summarization*. Examples of synchronic and diachronic relations are given in Table 1.

**Table 1.** Examples of synchronic and diachronic relations

| *Synchronic Relations* | *Diachronic Relations* | |
|---|---|---|
| – IDENTITY | – STABILITY | – VARIATION |
| – EQUIVALENCE | – ANTITHESIS | – IDENTITY |
| – ELABORATION | – POSITIVE GRADUATION | – ANALOGY |
| – CONTRADICTION | – NEGATIVE GRADUATION | |
| – PRECISENESS | | |

Each relation connects messages of the *same* type. In contrast to Radev's (2000) CST relations, our synchronic–diachronic relations are not dependent on the semantics of the sentences in order to be established, because they do not connect sentences. Instead, we have strict rules which connect certain messages according to the *values of their arguments*, which are predefined since they are taken from the ontology.

Things will become clearer with an example. Let us assume that we have the following two messages taken from the descriptions of two consecutive matches of the same team and from the same source:

```
performance (georgeas, general, round_17, excellent)
performance (georgeas, general, round_18, excellent)
```

What those messages state is that a certain player had, according to the author(s) of the articles, excellent performance in both the $17^{th}$ and $18^{th}$ round. According to our rules, given that the `value` and `entity` argument of this message are the same, we have a relation of type STABILITY connecting those two particular messages. If, on the other hand, we had the following two messages

```
performance (georgeas, general, round_17, excellent)
performance (georgeas, general, round_18, bad)
```

then the relation connecting them would be ANTITHESIS, because we have the same `entity` but *"contrasting"* `value`. Finally, if we had the messages

```
performance (georgeas, general, round_17, excellent)
performance (georgeas, general, round_18, mediocre)
```

then the resulting relation would be NEGATIVE GRADUATION since we have the same `entity` but *"close"* `values`. We have to note here that what is meant by *"contrasting"* and *"close"* `values` is something which is defined in the ontology, although here it is quite intuitively understood. Similarly, we have exact rules for each message, according to which the synchronic and diachronic relations are established. The rules take into account the values of the messages' arguments, as these are defined in the ontology.

But how does all that relate to summarization? As it has previously been stated, our system is query-based and relies on NLG for the production of the summary. The following example will explain how the above can be used for the creation of the summary. Let us assume that a user asks the question: *"What was the performance of Georgeas like during the first three rounds?"* In order to answer that question we analyze the query and we can see that we have to pinpoint the **performance** messages that are related to Georgeas for the first three rounds of the championship. For the sake of argument, let us assume that we only have three sources describing the matches of his team, and that we have already specified the messages and relations connecting them. In addition, the IE system has already identified the messages within the documents of our corpus. For each source, the query-specific messages are the following (what precedes each message is an identifier for it):

*Source 1:*
s1.1 **performance** (georgeas, general, round_1, excellent)
s1.2 **performance** (georgeas, general, round_2, excellent)
s1.3 **performance** (georgeas, general, round_3, mediocre)

*Source 2:*
s2.1 **performance** (georgeas, general, round_1, excellent)
s2.2 **performance** (georgeas, general, round_2, good)
s2.3 **performance** (georgeas, general, round_3, bad)

*Source 3:*
s3.1 **performance** (georgeas, general, round_1, excellent)
s3.3 **performance** (georgeas, general, round_2, excellent)
s3.3 **performance** (georgeas, general, round_3, bad)

Concerning the synchronic relations, for the first round all sources have exactly the same message, which means that there is a relation IDENTITY connecting those messages:

    IDENTITY(s1.1, s2.1)          IDENTITY(s1.1, s3.1)
    IDENTITY(s2.1, s3.1)

For the second and third round, not all sources agree, so the relations that exist are the following:

    IDENTITY(s1.2, s3.2)          CONTRADICTION(s1.2, s2.2)
    CONTRADICTION(s2.2, s3.2)     CONTRADICTION(s1.3, s2.3)
    IDENTITY(s2.3, s3.3)          CONTRADICTION(s1.3, s3.3)

Concerning the diachronic relations, the relations that hold are the following:

    NEGATIVE GRADUATION(s1.2, s1.3)     STABILITY(s3.1, s3.2)
    NEGATIVE GRADUATION(s2.1, s2.2)     STABILITY(s1.1, s1.2)
    NEGATIVE GRADUATION(s2.2, s2.3)
    NEGATIVE GRADUATION(s3.2, s3.3)

The above information can be used by the NLG system for the *content selection* phase. The NLG system, of course, will have to make more choices depending on several factors, such as the compression rate that the user wishes, etc. A candidate final summary can be the following:

*Georgeas's performance for the first two rounds of the championship was almost excellent. In the third round his performance deteriorated and was quite bad.*

A more concise summary could be:

*With the exception of the third round, Georgeas's performance was very good.*

From these examples the reader can glimpse the advantages that an abstractive summary has over an extractive one. The above summaries could not possibly have been created with an extractive approach since the generated sentences simply could not exist in the source documents. Furthermore, the manipulation of the relevant information for the creation of two different summaries, as happened above, cannot happen with an extractive approach. This means that, at least qualitatively, we achieve better results compared with an extractive system.

## 5   Conclusion

The aim of the paper is to propose a new approach for the specification and identification of cross-document relations which will enhance multi-document summarization.

    Currently the methodology is being applied to a specific topic, that of the descriptions of football matches, and has produced some promising results. We have to note though that not all of the stages of the methodology are fully automated yet; so far we have designed the architecture of the summarization system that we build, and collected its components. Our query-based summarization system involves an IE and an NLG system. These are currently being customized to the needs of our task. The ontology is currently being built using the CROSSMARC ontology management system.

    In the future we plan to examine the application of the methodology in other topics, in order for its strengths and weaknesses to be identified. For this

reason we plan to create the infrastructure needed for the adaptation of this methodology to other topics, *i.e.* to provide the infrastructure to support the creation of new entity types, message types and relation types.

## References

Karkaletsis, V., and C. D. Spyropoulos. 2003, November. "Cross-lingual Information Management from Web Pages." *Proceedings of the 9$^{th}$ Panhellenic Conference in Informatics (PCI-2003)*. Thessaloniki, Greece.

Mann, W. C., and S. A. Thompson. 1988. "Rhetorical Structure Theory: Towards a Functional Theory of Text Organization." *Text* 8 (3): 243–281.

Marcu, D. 2000. *The Theory and Practice of Discourse Parsing and Summarization*. The MIT Press.

Pazienza, M. T., A. Stellato, M. Vindigni, A. Valarakos, and V. Karkaletsis. 2003, June. "Ontology Integration in a Multilingual e-Retail System." *Proceedings of the Human Computer Interaction International (HCII'2003), Special Session on Ontologies and Multilinguality in User Interfaces*. Heraklion, Crete, Greece.

Radev, D. 2000, October. "A Common Theory of Information Fusion from Multiple Text Sources, Step One: Cross-Document Structure." *Proceedings of the 1st ACL SIGDIAL Workshop on Discourse and Dialogue*. Hong Kong.

Reiter, E., and R. Dale. 2000. *Building Natural Language Generation Systems*. Studies in Natural Language Processing. Cambridge University Press.

Zhang, Z., S. Blair-Goldensohn, and D. Radev. 2002. "Towards CST-Enhanced Summarization." *Proceedings of AAAI-2002*.